\documentclass[10pt,twocolumn,letterpaper]{article}

 \usepackage[pagenumbers]{cvpr} %

\usepackage{amsmath,amsfonts,bm}

\def\eqref#1{equation~\ref{#1}}

\def\1{\bm{1}}

\DeclareMathAlphabet{\mathsfit}{\encodingdefault}{\sfdefault}{m}{sl}
\SetMathAlphabet{\mathsfit}{bold}{\encodingdefault}{\sfdefault}{bx}{n}

\usepackage{bm}
\usepackage{multirow}
\usepackage{multicol}
\usepackage{pifont}
\usepackage{threeparttable}
\usepackage{wrapfig}
\usepackage{float}
\definecolor{cvprblue}{rgb}{0.21,0.49,0.74}
\usepackage[pagebackref,breaklinks,colorlinks,allcolors=cvprblue]{hyperref}

\def\paperID{*****}
\def\confName{CVPR}
\def\confYear{2026}

\title{
SM4RT: Learning Structured Motion Geometry for 4D Reconstruction
}

\author{
 Shing Ho J. Lin$^{*}$ \quad Wenzhao Zheng$^{*, \dagger}$ \quad Dong Zhuo$^{*}$ \quad Yuqi Wu \quad Jie Zhou \quad Jiwen Lu
 \vspace{2mm} \\
Intelligent Vision Group, Tsinghua University \\
Project Page: \url{https://wzzheng.net/SM4RT}\\
Code: \url{https://github.com/wzzheng/SM4RT}
}

\begin{document}

\twocolumn[{%
\maketitle
\vspace{-12mm}
\begin{figure}[H]
\hsize=\textwidth
\centering
\includegraphics[width=\textwidth]{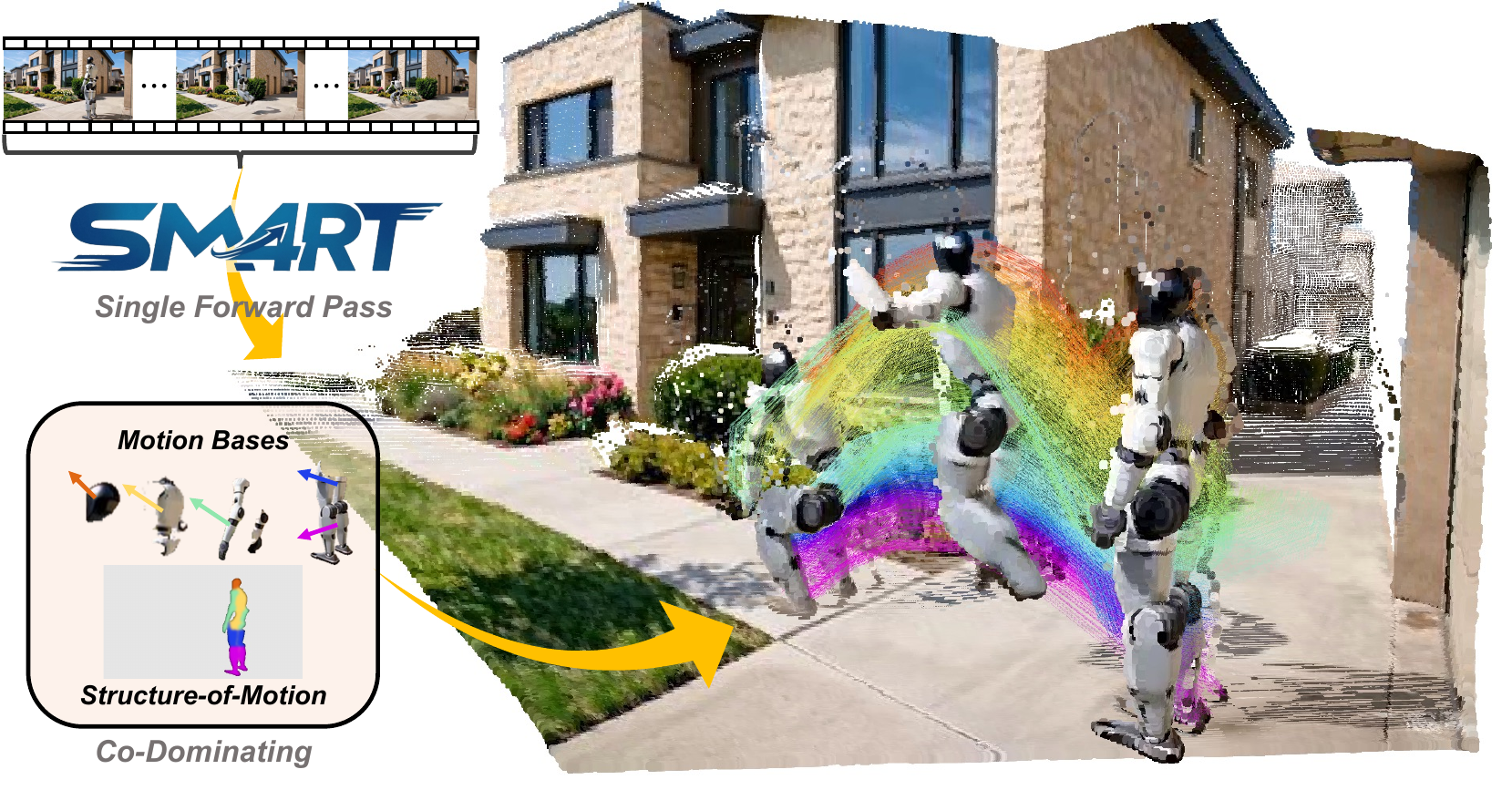}
\vspace{-9mm}
\caption{
\textbf{Structured motion geometry.}
\textbf{SM4RT} represents scene dynamics with \textit{Structure-of-Motion}: a set of motion bases to describe how each component evolves over time and time-shared point-to-base assignments to bind source points to motion bases.
They produce dense rigid transform maps that recover world-coordinate tracks and expose the structured motion geometry behind dynamic 4D scenes.
}
\label{fig:teaser}
\end{figure}
}]

\begingroup
\renewcommand\thefootnote{}
\footnotetext{
  $^*$Equal contributions.
  $^\dagger$Project leader.
}
\endgroup

\begin{abstract}
Geometry Foundation Models (GFMs) have substantially advanced monocular 3D reconstruction, yet extending this capability to 4D dynamic understanding remains a fundamental challenge.
Most existing motion perception methods (e.g., sparse tracking, dense point-wise flow) treat motion as independent point-wise displacements, ignoring the structured nature of physical motion.
However, real-world objects usually obey rigid-body kinematics, and points thus usually move collectively, not in isolation.
Motion itself possesses geometric structure: physical objects undergo a set of rigid-body transformations governed by $\mathrm{SE}(3)$, rather than unstructured point-wise displacements.
Building on this insight, we propose \textbf{SM4RT}, a \textbf{S}tructured \textbf{M}otion \textbf{4}D \textbf{R}econstruction \textbf{T}ransformer for end-to-end 3D reconstruction and structured motion perception.
SM4RT introduces \textit{Structure-of-Motion} (SoM) to represent scene dynamics, where scene motion is decomposed into a compact set of motion bases, each represented as a temporal sequence of 6D twists in $\mathfrak{se}(3)$.
Dense scene motion is then recovered by sparse, time-shared per-pixel assignment weights over these bases, ensuring points on the same object share a common rigid-body motion trajectory.
SM4RT introduces a parallel motion geometry encoder and decoder that jointly infer 3D geometry, world-coordinate motion, and scene kinematic structure in a single forward pass from monocular RGB video. 
SM4RT achieves strong motion reconstruction performance while preserving the geometric structure of scene motion. 
\end{abstract}

\section{Introduction}
Recent advances in Geometry Foundation Models (GFMs)~\cite{wang2025vggt,zhuo2025streaming,wu2025point3r,wang2025continuous,IVGT} have demonstrated remarkable capability in reconstructing 3D scene geometry from monocular video, facilitating applications from embodied intelligence~\cite{EmbodiedOcc} to autonomous driving~\cite{DVGT,DVGT2}.
With static geometry reconstruction now approaching practical fidelity, the next frontier is 4D dynamic understanding, which jointly infers scene structure and the underlying motion that governs its evolution.
Most recently, D4RT~\cite{zhang2026d4rt} has shown that a unified feed-forward transformer with a lightweight query decoder can make 4D reconstruction efficient and scalable, establishing queryable dynamic reconstruction as a strong paradigm.
Yet, such query-centric formulations primarily answer where points are across space and time and do not explicitly expose the structured motion geometry that explains why points move together.

Most existing motion perception methods are framed as point-wise trajectory estimation: either sparse point tracking~\cite{xiao2025spatialtrackerv2,wang2025scenetracker} or dense point-wise flow estimation~\cite{karhade2025any4d,ngo2025DELTAv2}.
Both treat motion as a collection of independent point-wise displacements, overlooking the relational dependencies among points.
While empirically effective for localization tasks, this formulation fails to capture the \textit{structured nature of physical motion}: accurate point-wise displacement does not directly imply meaningful dynamic understanding.
\textit{We reflect on how humans perceive motion.}
Humans perceive motion via sparse kinematic cues~\cite{johansson1973visual}: not at the pixel or point level, but at the level of objects and parts, enabling rapid and parsimonious motion understanding.
3D points that machines track are not atomic entities but samples drawn from continuous, physically grounded objects~\cite{RSCNN}.
Real-world objects usually obey rigid-body kinematics: points move \textit{collectively}, not in isolation.
Consider pure rotation: under a displacement view, rotating points appear as disorganized coordinate shifts, yet they share a single common motion governed by $\mathrm{SE}(3)$.
This motivates a shift in perspective: from \textit{geometry \textbf{with} motion} to the \textit{geometry \textbf{of} motion}, where motion itself possesses geometric structure~\cite{PersistentManifold} encoded in the symmetry groups of physical transformations, and any motion representation must respect these group-theoretic priors.

We propose \textbf{SM4RT}, a \textbf{S}tructured \textbf{M}otion \textbf{4}D \textbf{R}econstruction \textbf{T}ransformer framework that jointly infers 3D scene geometry, world-coordinate dense motion, and scene motion structure in a single forward pass given only monocular RGB video.
Rather than predicting independent point-wise displacements, SM4RT introduces \textit{Structure-of-Motion} (SoM): scene motion is decomposed into a compact set of $N$ motion bases, each represented as a temporal sequence of 6D twists in $\mathfrak{se}(3)$.
Dense scene motion is then recovered by combining these bases via sparse, time-shared per-pixel assignment weights, ensuring physically consistent motion throughout the scene.
We use the term \textit{Structure-of-Motion} to emphasize that the representation does not recover structure from motion as in classical SfM. Instead, it explicitly models the structure of motion itself, i.e., which points move together and how each coherent group evolves over time.
This twist-sequence decomposition is the key representational choice: the assignment map determines which points move together, while each motion base describes how the corresponding latent object or part moves over time.
Architecturally, SM4RT introduces a parallel Motion Geometry Encoder that distills dynamic cues from intermediate geometry features and a Motion Geometry Decoder that jointly predicts the motion bases and per-pixel assignment map.
Entropy regularization and a pseudo-background constraint further encourage parsimonious, physically meaningful decompositions.
Through shared kinematic parameters and implicit entity binding, SM4RT enforces global motion coherence, preserves topological integrity over time, and yields a representation that is both physically interpretable and compositionally structured.
Beyond point tracking, the twist representation unlocks structured motion interpolation and future state prediction grounded in rigid-body kinematics, capabilities that are out of reach for point-wise displacement methods.
SM4RT achieves state-of-the-art performance on existing motion perception benchmarks, demonstrating that structured motion perception and accurate point-wise trajectory estimation are mutually reinforcing rather than conflicting objectives.

\section{Related Work}

\textbf{Geometry Foundation Models.}
Building on the success of large-scale vision transformers~\cite{dosovitskiy2020image,oquabdinov2}, Geometry Foundation Models (GFMs) have recently emerged as feed-forward 3D reconstruction systems that avoid per-scene optimization and learn generalizable geometric priors from large-scale data.
These models~\cite{wang2025vggt,wang2025continuous,zhuo2025streaming,wu2025point3r} share a common paradigm: given a set of images or video frames, they predict camera parameters, depth maps, and point maps in a single forward pass.
VGGT~\cite{wang2025vggt} establishes a unified backbone for multi-view geometry, jointly reasoning about appearance and 3D structure across frames.
CUT3R~\cite{wang2025continuous} maintains a persistent scene state for efficient online reconstruction, while StreamVGGT~\cite{zhuo2025streaming} extends this to streaming video with incremental frame processing.
Point3R~\cite{wu2025point3r} revisits point-centric representations to improve robustness under challenging viewpoint changes.
Beyond scene-level reconstruction, MapAnything~\cite{keetha2026mapanything} targets scale-aware mapping, and IVGT~\cite{IVGT} explores implicit neural scene representations for 3D mesh construction.
SM4RT extends this foundation from static geometry to 4D motion.

\textbf{Motion Perception.}
Motion perception spans a broad spectrum from 2D pixel tracking to fully 3D dense estimation, yet most methods represente motion as point-wise displacement between frames.
In 2D, a large body of work~\cite{doersch2023tapir,CoTracker3,lai2026cowtracker,karaev2023cotracker} has advanced long-range point tracking through temporal consistency modeling, sliding-window attention, and feature warping, achieving robust performance on occluded and fast-moving targets.
Lifting to 3D, methods such as SceneTracker~\cite{wang2025scenetracker} and SpatialTrackerV2~\cite{xiao2025spatialtrackerv2} incorporate monocular depth priors and geometric scene priors to track sparse 3D queries, though they remain limited to sparse inputs by design.
For dense 3D motion, recent methods~\cite{ngo2025delta,ngo2025DELTAv2,feng2025st4rtrack,liu2025traceanythingrepresentingvideo,karhade2025any4d,sucar2026v,luo20264rc} push toward per-pixel world-coordinate tracking from monocular video.
Notably, St4RTrack~\cite{feng2025st4rtrack} maps pixel motion into a global 4D coordinate space, TraceAnything~\cite{liu2025traceanythingrepresentingvideo} fits parametric B-spline trajectories for smooth motion fields, and 4RC~\cite{luo20264rc} recovers geometry and dense trajectories through a unified query-response mechanism.
Recently, D4RT~\cite{zhang2026d4rt} introduces a unified feed-forward transformer with a lightweight query decoder, enabling efficient reconstruction by flexibly probing 3D point positions across source time, target time, and camera coordinates.
However, they produce independent point-wise motion estimates without any notion of object-level structure, making them fundamentally limited for structured dynamic scene understanding.
Rather than focusing only on efficient point-time querying, SM4RT models motion with rigid kinematic entities and exposes object/part-level motion structure.

\textbf{Motion Decomposition.}
Some prior work also recognizes that scene motion can be decomposed into a compact set of dominant rigid components~\cite{teed2021raft,wang2025shape,zhou2025motion3d}, echoing classical multi-body motion segmentation~\cite{vidal2008multiframe}.
RAFT-3D~\cite{teed2021raft} predicts a dense $\mathrm{SE}(3)$ field for image pairs, but requires RGBD input and does not generalize to monocular RGB video.
Motion4D~\cite{zhou2025motion3d} alternates between motion and semantic optimization with the aid of SAM2 masks.
Shape of Motion~\cite{wang2025shape} is closely related to our representation: it exploits the low-dimensional structure of dynamic scenes through compact $\mathrm{SE}(3)$ motion bases and soft point-to-basis decomposition.
However, it recovers this structure through per-scene optimization with external priors such as monocular depth and long-range 2D tracks.
In contrast, SM4RT learns an amortized Structure-of-Motion representation inside a feed-forward Geometry Foundation Model, directly predicting time-shared assignments and temporal twist-sequence motion bases from RGB video in a single forward pass.
They demonstrate the viability of structured motion decomposition but share key limitations: they rely on auxiliary inputs such as depth or segmentation masks, require iterative test-time optimization, and do not learn a feed-forward motion representation from data.
SM4RT addresses these limitations by learning Structure-of-Motion in a single forward pass from monocular RGB video, with no auxiliary inputs and no test-time optimization.

\section{Proposed Approach}

\subsection{Geometry with Motion to Geometry of Motion}

\textbf{Geometry Foundation Models.}
Given a monocular RGB video $\bm{I}$ of $S$ frames, a GFM $\mathcal{F}_\text{geo}$ infers per-frame camera intrinsics $\mathbf{K}$, extrinsics $[\mathbf{R}\mid\mathbf{t}]$, and depth maps $\bm{D}$ in a single forward pass:
\begin{equation}
    (\bm{D},\, \mathbf{K},\, [\mathbf{R}\mid\mathbf{t}]) = \mathcal{F}_\text{geo}(\bm{I}).
\end{equation}
These outputs unproject to a world-coordinate point cloud $\bm{P}$ for each frame, constituting a complete static 3D representation of the scene.

\begin{figure}[t]
    \centering
    \includegraphics[width=\linewidth]{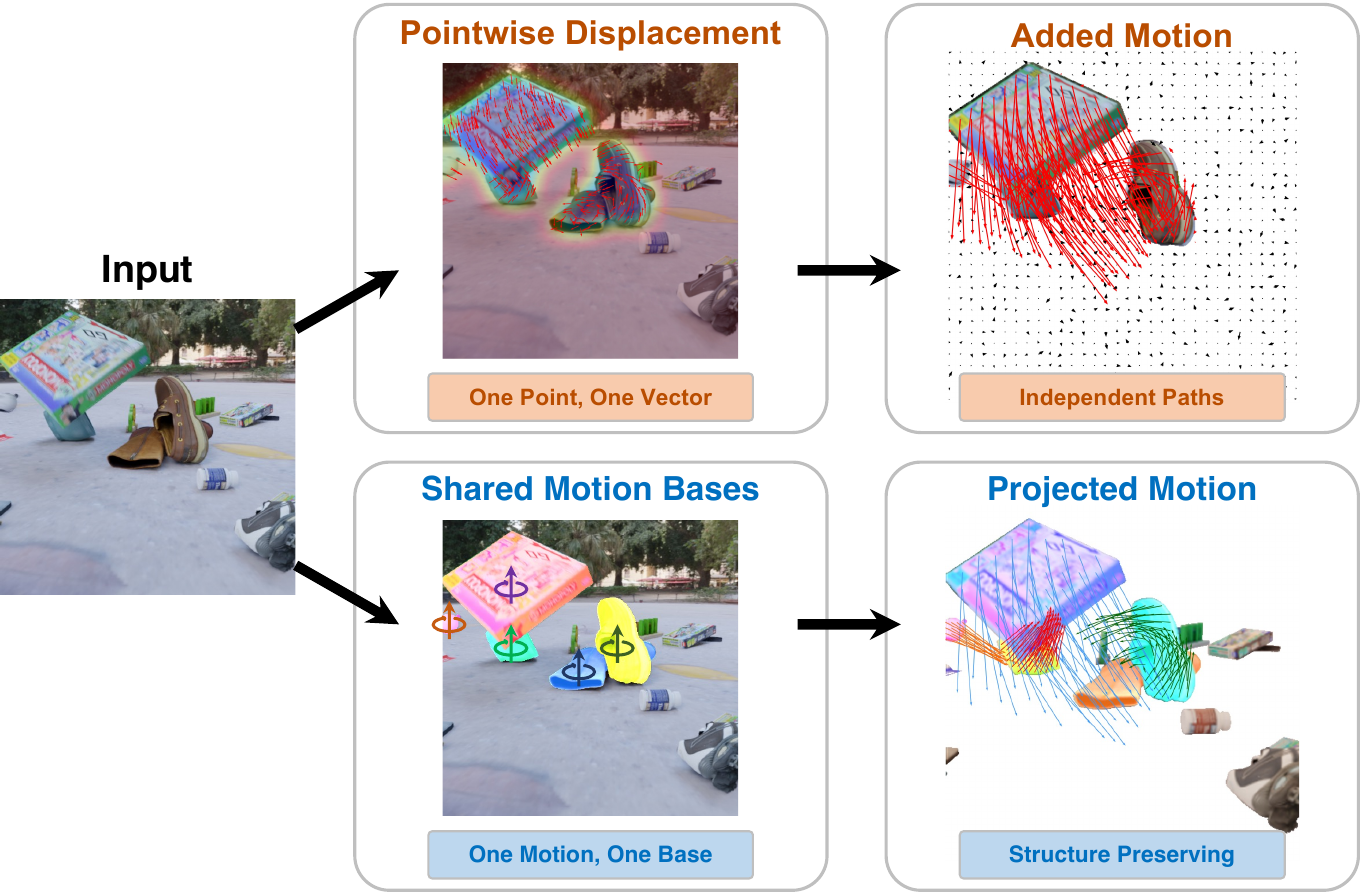}
    \vspace{-7mm}
    \caption{\textbf{Comparison of motion representations.} The proposed representation preserves the structure of object motion, providing a more structured description than standard per-pixel displacement and encouraging intra-object homogeneity.}
    \label{fig:TwistRepresentation}
    \vspace{-5mm}
\end{figure}

\textbf{Existing Motion Perception.}
Existing methods~\cite{karhade2025any4d,sucar2026v,luo20264rc} usually formulate motion as point-wise trajectory estimation. Let $t=0$ denote the source frame and $u$ denote a source-frame position. Dense methods predict a displacement field $\bm{\Delta}$ over all target frames:
\begin{equation}
    \hat{\bm{p}}_{u,t} = \bm{p}_{u,0} + \bm{\Delta}_{u,t}, \quad t = 1, \ldots, S.
\end{equation}
Each pixel $u$ is predicted independently with no self-constrained structural constraint coupling it to its neighbors.
In the sparse case, a subset of query points $\{\bm{q}_k\}_{k=1}^K$ is tracked individually as $\hat{\bm{p}}_{k,t} = f_\theta(\bm{I}, \bm{q}_k)$, again not exploiting inter-point relations.

\textbf{The Structural Gap.}
This independent displacement formulation places no constraint between pixels of the same physical object.
Yet for any rigid body undergoing transformation $\mathbf{T} \in \mathrm{SE}(3)$, all points on the object satisfy the coupling constraint:
\begin{equation}
    \bm{p}_{u,t} = \mathbf{T}_t\, \bm{p}_{u,0}, \quad \forall u \in \text{object}.
\end{equation}
A scene with $N$ rigid objects thus has only $6N$ motion degrees of freedom per target frame, far fewer than the $3HW$ degrees assumed by point-wise displacement.
Ignoring this structure forces the model to independently fit $HW$ unconstrained displacements, which is a severely over-parameterized and physically inconsistent formulation.

\textbf{Our Paradigm.}
As shown in \Cref{fig:TwistRepresentation}, we propose to model scene motion as a sparse mixture of $N$ motion bases.
Each pixel $u$ is assigned to the bases via time-shared weights $\bm{w}_u = [w_{u,1}, \ldots, w_{u,N}]^\top$, $\sum_i w_{u,i} = 1$.
Each motion base is a temporal twist sequence $\{\bm{\xi}_{i,t}\}_{t=1}^{S}$, where $\bm{\xi}_{i,t}\in\mathfrak{se}(3)$ describes the rigid motion of basis $i$ at target frame $t$.
The dense scene motion at pixel $u$ and frame $t$ is:
\begin{equation}
    \bm{\Xi}_{u,t} = \sum_{i=1}^{N} w_{u,i}\, \bm{\xi}_{i,t} \in \mathfrak{se}(3), \qquad
    \bm{p}_{u,t} = \mathrm{Exp}(\hat{\bm{\Xi}}_{u,t})\, \bm{p}_{u,0}.
    \label{eq:paradigm}
\end{equation}
We omit homogeneous coordinates for clarity.
Points sharing the same dominant basis follow the same temporal rigid-motion trajectory, implicitly recovering object-level kinematic structure.
In other words, $\bm{W}$ answers \textit{which points move together}, while $\{\bm{\xi}_{i,t}\}_{t=1}^{S}$ answers \textit{how each group moves over time}.
Concretely, instead of predicting point-wise locations with $3SHW$ DoF, we predict a single assignment map and temporal motion bases with $NHW + 6SN$ DoF. Since $N \ll HW$ and assignments are sparse, this representation is both parsimonious and physically grounded, with only weak scaling in the number of frames.

\subsection{Structured Motion Geometry}

We define the geometry of motion as the structured organization of scene dynamics induced by shared physical transformations.
This is the motivation behind our \textit{Structure-of-Motion} representation: rather than treating motion as only a set of point trajectories, SoM factorizes motion into time-shared point-to-base assignments and temporal twist-sequence motion bases. 
Rather than treating motion as independent point-wise displacement, SM4RT models groups of points as belonging to coherent kinematic entities governed by rigid-body transformations in $\mathrm{SE}(3)$, with their Lie algebra generators $\mathfrak{se}(3)$ serving as motion bases.
This perspective captures object-level motion coherence, topology preservation, and physical consistency, providing an interpretable bridge between dense 4D reconstruction and high-level dynamic scene structure understanding.

\textbf{Twist Representation.}
A rigid body motion can be described by how it translates and how it rotates. 
We use a \textit{twist} for this purpose: a compact 6D motion vector $\bm{\xi} = [\bm{\rho}, \bm{\phi}] \in \mathbb{R}^6$, where $\bm{\rho} \in \mathbb{R}^3$ denotes the translational component and $\bm{\phi} \in \mathbb{R}^3$ denotes the rotational component.
In other words, one twist specifies one rigid-body motion pattern shared by a group of points. 
Formally, the twist $\bm{\xi}$ corresponds to an element of the Lie algebra $\mathfrak{se}(3)$ via:
\begin{equation}
\hat{\bm{\xi}} =
\begin{bmatrix}
[\bm{\phi}]_\times & \bm{\rho} \\
\mathbf{0}^\top & 0
\end{bmatrix} \in \mathfrak{se}(3),
\end{equation}
where $[\bm{\phi}]_\times \in \mathfrak{so}(3)$ is the skew-symmetric matrix of $\bm{\phi}$. 
The exponential map converts this 6D motion description into a finite rigid transformation in $\mathrm{SE}(3)$~\cite{murray1994mathematical}, which maps a source point $\bm{p}$ to a target point $\bm{p}'$:
\begin{equation}
\bm{p}' = \mathrm{Exp}(\hat{\bm{\xi}}) \bm{p},
\end{equation}
where homogeneous coordinates are omitted.

\begin{figure}[!t]
    \centering
    \includegraphics[width=\linewidth]{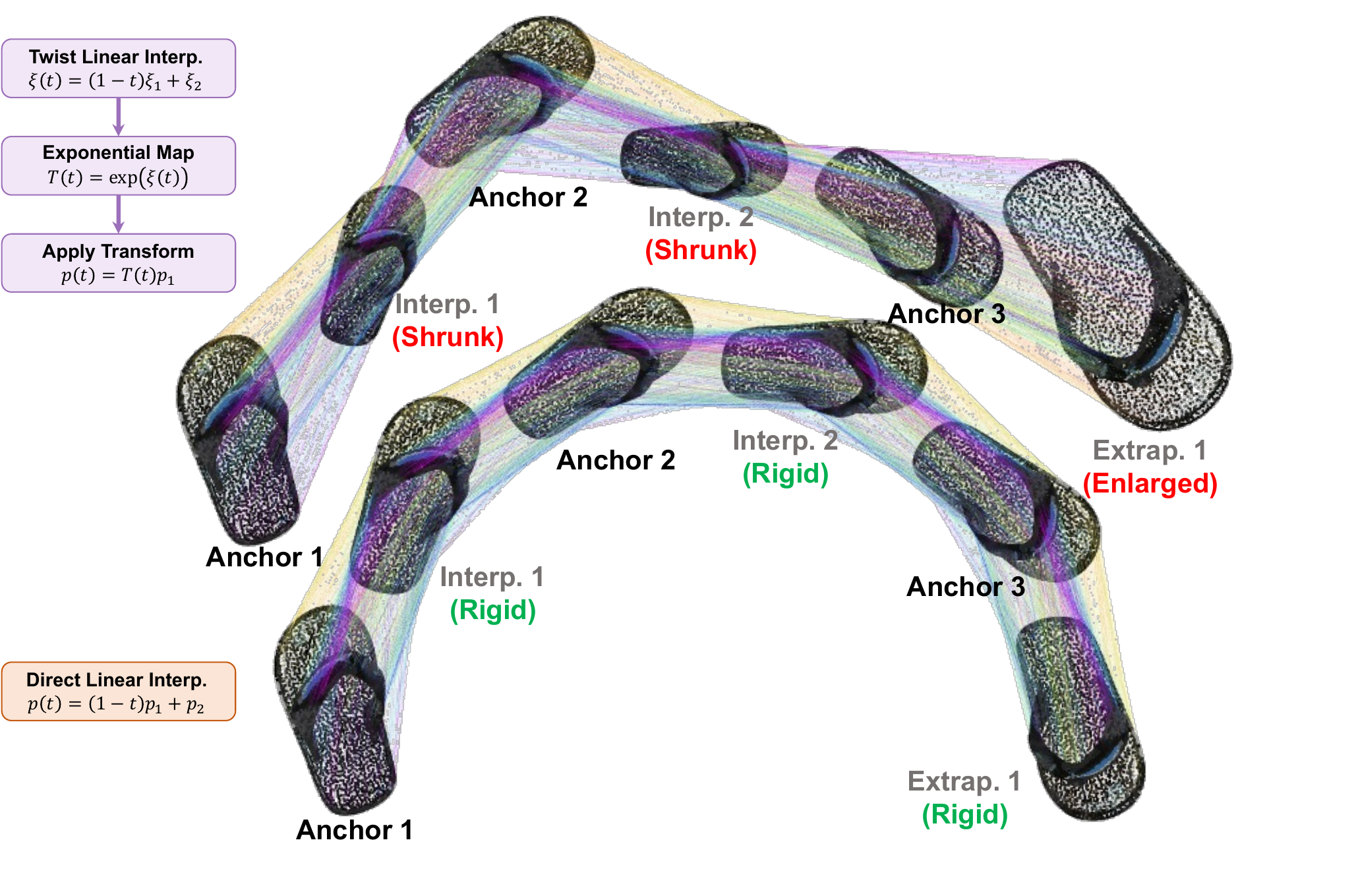}
    \vspace{-10mm}
    \caption{
    \textbf{Comparison of parameter-free linear interpolation and extrapolation in base space (top) and track space (bottom).} Using anchor frames, interpolation generates the intermediate state and extrapolation predicts the future state. Operating in base space preserves structural integrity, while track-space results exhibit deformation. The two are equivalent for objects undergoing pure translation.}
    \label{fig:motion_interp}
    \vspace{-5mm}
\end{figure}

\begin{figure*}[!t]
    \centering
    \includegraphics[width=\linewidth]{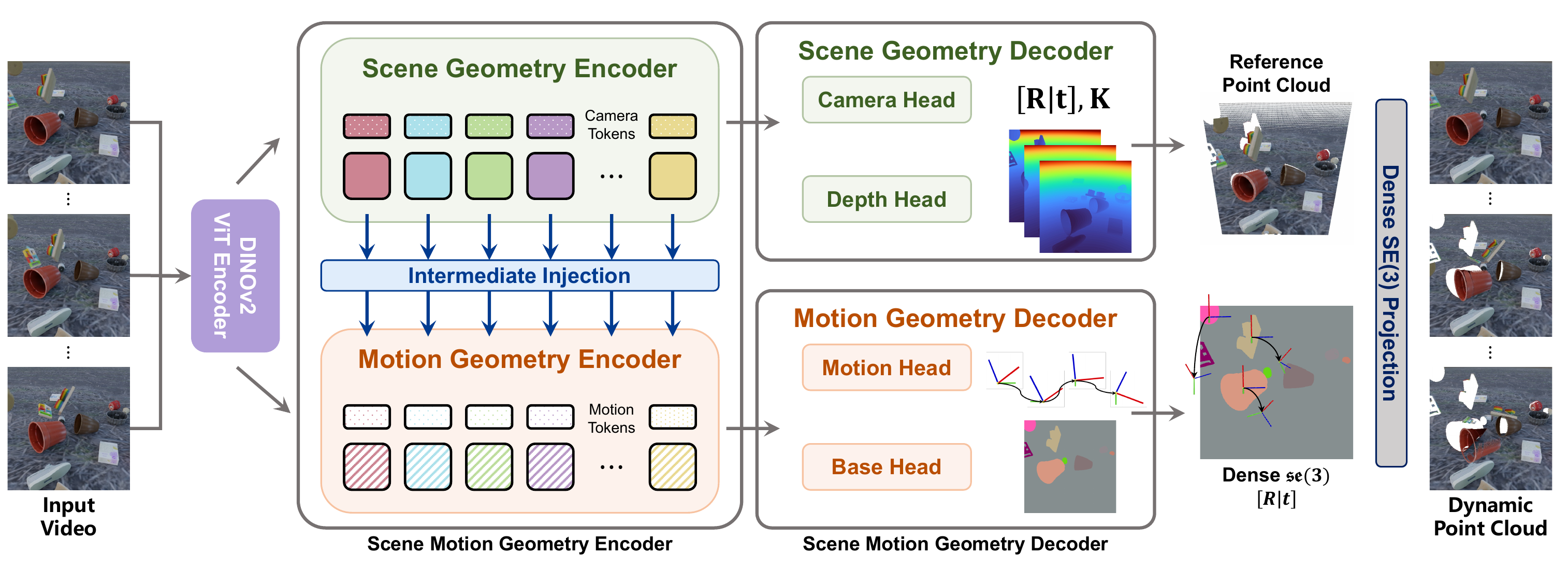}
    \vspace{-9mm}
    \caption{\textbf{Overview of our SM4RT.} Given a monocular input video, SM4RT extracts latent geometry tokens with a DINOv2 backbone, then decouples them into scene geometry and motion geometry tokens. The scene decoder predicts camera parameters and depth, while the motion decoder estimates sparse $\mathfrak{se}(3)$ transforms and motion base weights for dense SE(3) projection.}
    \label{fig:SM4RT}
    \vspace{-5mm}
\end{figure*}

\textbf{Structure-of-Motion via Motion Bases.}
We treat scene motion as a linear composition of $N$ shared motion bases in twist space.
By decomposing motion into linear and angular components, twist encodes object movement as a unified SE(3) transformation, overcoming the heterogeneity of dense displacement maps.
Concretely, for a video $\bm{I}$, we predict time-shared assignment weights $\bm{W}=\{w_{u,i}\}$ and $N$ temporal motion bases $\{\bm{\xi}_{i,t}\}$ separately.
The assignment map $\bm{W}$ is shared across target frames, binding each source point to latent kinematic entities, while each motion base provides a sequence of twists describing the entity's motion over time.
They are combined via weighted sum in the twist tangent space to obtain the dense twist $\bm{\Xi}_{u,t}$ and point-wise transform $\mathbf{T}_{u,t}$:
\begin{equation}
    \bm{\Xi}_{u,t}=\sum_{i=1}^{N} w_{u,i}\bm{\xi}_{i,t}, \qquad
    \mathbf{T}_{u,t}=\mathrm{Exp}(\hat{\bm{\Xi}}_{u,t}).
\end{equation}
The final 3D point positions are obtained by applying the transforms to source-frame points:
\begin{equation}
    \bm{p}_{u,t} = \mathbf{T}_{u,t}\bm{p}_{u,0}.
\end{equation}
For deformable objects such as animals or human bodies, this representation remains applicable, as deformable motion can be expressed as a linear combination of neighboring skeletal rigid motions.

Our SoM replaces the over-parameterized per-pixel displacement field with a compact, physically grounded decomposition, reducing the effective motion degrees of freedom from $3SHW$ to $NHW+6SN$ while enforcing object-level coherence by construction.
The key challenge is then to learn the temporal motion bases $\{\bm{\xi}_{i,t}\}$ and per-pixel assignments $\bm{W}$ end-to-end from RGB video alone, which the architecture described next is designed to address.

\textbf{Emerging Capability of Structured Motion.}
Operating in twist space $\mathfrak{se}(3)$ unlocks structured motion interpolation and extrapolation.
Existing parameter-free methods~\cite{liu2025traceanythingrepresentingvideo} can only perform point-wise extrapolation via linear tangent interpolation or cubic B-splines, which breaks object structure or causes severe jittering. 
In contrast, because SM4RT models the coherent motion of entire kinematic entities rather than individual points, interpolation in $\mathfrak{se}(3)$ naturally respects rigid-body motion and preserves geometric consistency across the sequence.

As shown in \Cref{fig:motion_interp}, operating in motion-base space produces more coherent intermediate and future states, since points assigned to the same kinematic component move consistently. This demonstrates that our Structure-of-Motion representation provides a compact and physically grounded basis for temporal motion reasoning. 
However, like existing parameter-free prediction methods, SM4RT cannot explicitly handle complex physical interactions such as collisions, rebounds, or contact dynamics.

\subsection{SM4RT: Structured Motion 4D Reconstruction Transformer}

SM4RT processes an input video $\bm{I}$ through two parallel branches: a \textit{Scene Geometry} branch that recovers 3D structure and camera parameters, and a \textit{Motion Geometry} branch that infers the motion bases and per-pixel assignments jointly.
The two branches share a common geometry-aware backbone and interact through cross-attention, allowing motion reasoning to be grounded in 3D scene structure.
The overall architecture is illustrated in~\Cref{fig:SM4RT}.

\textbf{Scene Geometry Encoder.}
Each frame $\bm{I}_{t}$ is patchified and encoded via a pretrained DINOv2~\cite{oquabdinov2} backbone into patch features $\bm{F}$.
A learnable camera token and register tokens are appended per frame, and the full sequence is fed into a geometry-aware backbone, a pretrained DepthAnythingV3~\cite{lin2026depth} with local-global attention, to produce geometry tokens $\bm{G}$ enriched with 3D geometric cues, epipolar geometry, and cross-frame consistency.

\textbf{Scene Geometry Decoder.}
The Scene Geometry Decoder comprises a \textit{Camera Head} and a \textit{Depth Head}, both adopted from DepthAnythingV3~\cite{lin2026depth}.
The Camera Head outputs per-frame extrinsics and intrinsics. The Depth Head outputs depth maps $\bm{D}$.
Together, they yield anchored world-coordinate source points via unprojection:
\begin{equation}
    \bm{P}^{0} = \mathrm{Reproject}(\bm{D}^{0}, \mathbf{K}^{0}),
\end{equation}
where world coordinates are defined in the camera coordinate system of the source frame.

\textbf{Motion Geometry Encoder.}
While the geometry-aware backbone encodes rich scene structure, it lacks motion-specific inter-frame cues.
We therefore introduce a parallel Motion Geometry Encoder that uses four intermediate geometry token layers $\left\{\bm{G}'_{i}\mid i=0,1,2,3\right\}$ as auxiliary references.
$N$ motion tokens $\bm{m}$, initialized with per-frame time embeddings, first attend to $\bm{G}_{0}$ via cross-attention to obtain $\bm{m}'$, which is then concatenated as $[\bm{m}', \bm{G}_{0}']$ and fed through 4 layers of frame motion attention with positional embeddings.
After each layer, the corresponding intermediate geometry token is injected via a projection layer as a structural reference.
This design allows motion tokens to aggregate dynamic cues from the geometry stream while capturing long-range temporal dependencies.

\textbf{Motion Geometry Decoder.}
The Motion Geometry Decoder recovers physical motion from the motion tokens through two heads.
The \textit{Base Head} applies a Dense Prediction Transformer (DPT)~\cite{ranftl2021vision} to produce per-pixel base assignment weights $\bm{W}\in\mathbb{R}^{H \times W \times N}$. $\mathrm{SparseMax}(\cdot)$~\cite{sparsemax} is applied along $N$ to induce sparsity and promote hard assignments.
The \textit{Motion Head} applies an MLP to decode motion tokens $\bm{m}$ into temporal twist sequences $\{\bm{\xi}_{i,t}\}$.
The dense twist $\bm{\Xi}_{u,t}$ is then obtained by weighting the temporal bases with $\bm{W}$, followed by the exponential map to obtain point-wise rigid transforms $\mathbf{T}_{u,t}$.
The two branches jointly produce the full set of outputs defined in~\Cref{eq:paradigm}: source-frame world points $\bm{P}^0$ from the Scene Geometry branch, and temporal motion bases $\{\bm{\xi}_{i,t}\}$ with assignment weights $\bm{W}$ from the Motion Geometry branch, in a single forward pass without any test-time optimization.

\textbf{Post Processing (Optional).}
To enhance structured representation, we include an adaptive grouping strategy used in segmentation~\cite{li2026iggt} denoted as \textit{SM4RT (Adaptive)}. 
Upon obtaining $\bm{W}$, hdbscan~\cite{campello2013density} (implemented by \texttt{cuml}) is applied to adaptively determine clustering classes and masks. The base of each masked region is replaced with the average value of its assigned region. 
The updated $\bm{W}'$ is then fed into~\Cref{eq:paradigm} and subsequent steps.

\subsection{Training}

The total training objective combines geometry supervision, motion point supervision, and motion base regularization:
\begin{equation}
    \mathcal{L} = \lambda_1 \mathcal{L}_{\textit{geo}} + \lambda_2 \mathcal{L}_{\textit{point}} + \lambda_3 \mathcal{L}_{\textit{base}},
\end{equation}
where $\lambda$ are balancing weights.

\textbf{Geometry Supervision.} We supervise the predicted depth $\bm{D}$, camera pose $[\mathbf{R}\mid\mathbf{t}]$, and scene scale. 
The total loss is defined as $\mathcal{L}_{\text{geo}} = \mathcal{L}_{\text{depth}} + \mathcal{L}_{\text{cam}} + \mathcal{L}_{\text{scale}}$. Specifically, we employ a normalized $\ell_1$ loss for depth estimation over valid pixels $\mathcal{M}$:
\begin{equation}
\mathcal{L}_{\textit{depth}} = \frac{1}{|\mathcal{M}|} \sum_{i \in \mathcal{M}} \left| \text{norm}(\hat{\bm{D}}_i) - \text{norm}(\bm{D}^*_i) \right|.
\end{equation}
For camera parameters, we apply a standard $\ell_1$ loss against the ground truth. To constrain the inherent scale ambiguity, we compute the scale loss by minimizing the log-difference between the mean Euclidean distance of source-frame points $\bm{P}^0$ and a unit scale:
\begin{equation}
\mathcal{L}_{\textit{scale}} = \left( \log \left( \text{mean}(|\bm{P}^0|_2) \right) - \log(1.0) \right)^2.
\end{equation}

\textbf{Motion Point Supervision.}
Datasets with dense annotations allow foreground and background pixels to be decoupled via mask to avoid domination by static points, whereas sparse tracking datasets lack exhaustive per-pixel labels and do not require such decoupling. 
We therefore apply different supervision strategies for $\mathcal{L}_{\textit{point}}$ depending on data type:
\begin{equation}
\begin{aligned}
&\mathcal{L}_{\textit{dense}} = \mathcal{L}_{\textit{fg}} + \alpha \cdot\mathcal{L}_{\textit{bg}} \\
&= \frac{1}{|\mathcal{M}|}\sum_{u \in \mathcal{M}} \ell( \mathbf{p}_u, \mathbf{p}'_u )
+
\frac{\alpha}{|\neg\mathcal{M}|}\sum_{u \in \neg\mathcal{M}} \ell( \mathbf{p}_u, \mathbf{0}), \\
&\mathcal{L}_{\textit{sparse}} = \sum_{u \in \mathcal{M}} \ell( \mathbf{p}_u, \mathbf{p}'_u ),
\end{aligned}
\end{equation}
where $\mathcal{M} \in \{0,1\}$ is the selected mask, $\mathbf{p}_u$ and $\mathbf{p}'_u$ are the deviatoric positions from source to target of prediction and ground-truth at pixel $u$, and $\alpha$ downweights the background. 

\textbf{Motion Base Supervision.}
Multiple auxiliary losses regularize the base assignment map $\bm{W}$ to encourage physically meaningful motion assignment.

\textit{Entropy regularization} promotes sparse, unambiguous assignments by penalizing diffuse distributions over bases:
\begin{equation}
    \mathcal{L}_{\textit{entropy}} = -\sum \bm{W} \odot \log(\bm{W}).
\end{equation}

\textit{Pseudo-background regularization} addresses the tendency of the static background base to bleed into dynamic regions. Since the background occupies the largest spatial area in most scenes, we identify the pseudo-background (PB) base as the mode of the dominant assignment indices, and penalize its assignment in non-background regions:
\begin{equation}
\begin{aligned}
\mathcal{L}_{\textit{pseudo}} &= \left\|
\textbf{1}_{(\arg\max_N \bm{W})\neq n_{\textit{pseudo}}}
\odot (\bm{W}_{n_{\textit{pseudo}}}) \right\|_2^2, \\
n_{\textit{pseudo}} &= \text{mode}(\arg\max_N \bm{W}).
\end{aligned}
\end{equation}

\textit{Twist singularity regularization.}
A dense SE(3) motion field can collapse into translation-only motion, weakening the uniform-motion property. 
To tackle this, we introduce a translation supervision parallel to point supervision. 
\begin{equation}
    \mathcal{L}_{\textit{singular}} = \frac{1}{|\mathcal{M}|}\sum_{u \in \mathcal{M}} \ell( \mathbf{t}_u, \mathbf{t}'_u ),
    \label{eq:singular}
\end{equation}
where $\mathbf{t}_{u}$ and $\mathbf{t}'_{u}$ are the translation components of the predicted and ground-truth dense SE(3) transforms at pixel $u$. 
This term is used for samples with dense SE(3) annotations. 

\begin{table*}[!t]
\centering
\caption{\textbf{Performance on video depth estimation.} We follow the protocols of~\cite{zhang2025monst3r}, reporting AbsRel ($\downarrow$) and $\delta_{<1.25} (\uparrow)$.}
    \vspace{-3mm}
\label{tab:depth}
\resizebox{\linewidth}{!}{
\begin{tabular}{l|cccccccccc}
\toprule
 & 
 \multicolumn{2}{c}{Bonn} & 
 \multicolumn{2}{c}{Sintel} & 
 \multicolumn{2}{c}{KITTI} & 
 \multicolumn{2}{c}{7Scenes} & 
 \multicolumn{2}{c}{ScanNetV2} \\
\midrule
Method & 
AbsRel $\downarrow$ & $\delta_{<1.25} \uparrow$ & 
AbsRel $\downarrow$ & $\delta_{<1.25} \uparrow$ & 
AbsRel $\downarrow$ & $\delta_{<1.25} \uparrow$ & 
AbsRel $\downarrow$ & $\delta_{<1.25} \uparrow$ & 
AbsRel $\downarrow$ & $\delta_{<1.25} \uparrow$ 
\\
\midrule
MonST3R~\cite{zhang2025monst3r} & 
0.072 & 95.48 & 
0.309 & 54.23 & 
0.095 & 91.59 & 
0.132 & 84.57 &
0.092 & 92.25
\\
MapAnythingV1~\cite{keetha2026mapanything} & 
0.068 & 96.80 &
0.275 & 59.58 &
0.066 & 95.81 &
0.076 & 92.82 &
0.039 & 98.34 \\
VGGT~\cite{wang2025vggt}    & 
\underline{0.051} & 97.18 & 
0.306 & 62.68 & 
\textbf{0.051} & 96.88 & 
0.073 & 91.79 & 
\textbf{0.023} & \underline{98.84} \\
DepthAnythingV3~\cite{lin2026depth}     & 
\textbf{0.046} & \underline{97.38} & 
\underline{0.186} & \underline{70.66} & 
\underline{0.052} & \textbf{97.35} & 
\textbf{0.066} & \underline{92.74} & 
0.032 & 98.04 \\
\midrule
SM4RT & 
0.054 & \textbf{97.49} & 
\textbf{0.169} & \textbf{75.51} & 
0.056 & \underline{97.20} & 
\underline{0.067} & \textbf{92.91} & 
\underline{0.029} & \textbf{98.86} \\
\bottomrule
\end{tabular}
}
\vspace{-3mm}
\end{table*}

\begin{table*}[!t]
  \centering
  \caption{\textbf{Performance on 3D reconstruction.} We adopt DA3Bench~\cite{lin2026depth}, reporting performance on Pose Estimation (Pose) and Unposed Reconstruction (Rec.). For Pose, we report Auc3 / Auc30 ($\uparrow$). For Rec., we report CD $(\downarrow)$ for DTU, and F1 Score $(\uparrow)$ for the rest.}
    \vspace{-3mm}
  \label{tab:DepthAnythingV3recon}
\resizebox{\linewidth}{!}{
  \begin{tabular}{l|rr|rr|rr|rr|rr}
    \toprule
    &
    \multicolumn{2}{c|}{\textbf{HiRoom}   } & 
    \multicolumn{2}{c|}{\textbf{ETH3D}    } & 
    \multicolumn{2}{c|}{\textbf{DTU}      } & 
    \multicolumn{2}{c|}{\textbf{7Scenes}  } & 
    \multicolumn{2}{c }{\textbf{ScanNet++}} \\
    \midrule
    \textbf{Model} & 
    \multicolumn{1}{c}{Pose} & \multicolumn{1}{c|}{Rec.} &
    \multicolumn{1}{c}{Pose} & \multicolumn{1}{c|}{Rec.} &
    \multicolumn{1}{c}{Pose} & \multicolumn{1}{c|}{Rec.} &
    \multicolumn{1}{c}{Pose} & \multicolumn{1}{c|}{Rec.} &
    \multicolumn{1}{c}{Pose} & \multicolumn{1}{c}{Rec.}
    \\
    \midrule
    MapAnythingV1~\cite{keetha2026mapanything} & 
    22.13 / 83.51 & 27.48 &
    21.96 / 79.36 & 44.89 &
     5.56 / 77.89 & 7.469 &
    18.36 / 81.24 & 39.46 &
    27.75 / 87.59 & 34.62 \\
    Pi3  & 
    66.88 / 94.78 & 67.68 & 
    35.30 / 87.30 & 71.61 & 
    62.54 / 94.82 & 3.388 & 
    26.34 / 86.18 & 43.43 & 
    50.36 / 92.42 & 58.54 \\
    VGGT~\cite{wang2025vggt} & 
    49.03 / 87.96 & 57.20 & 
    26.33 / 80.80 & 55.95 & 
    79.19 / 85.27 & 2.030 & 
    23.63 / 84.91 & 47.14 & 
    59.48 / 94.43 & 66.98 \\
    DepthAnythingV3~\cite{lin2026depth} & 
               79.39  /            95.26  & 86.08             & 
    \textbf{47.84}    / \underline{90.81} & \textbf{78.69}    & 
    \underline{93.25} / \underline{99.30} & 1.819            & 
    \underline{27.46} / \underline{86.68} & 49.85             & 
    \textbf{83.90}    /    \textbf{98.17} & \underline{78.30} \\
    \midrule
    Any4D~\cite{karhade2025any4d} & 
    11.22 / 78.87 & 15.80 & 
     4.63 / 61.46 & 28.19 & 
     2.15 / 74.04 & 6.206 & 
    11.16 / 78.32 & 28.38 & 
     6.43 / 74.62 & 24.35 \\
    VDPM~\cite{sucar2026v} &
    28.15 / 85.45 & 41.12 &
    11.41 / 73.99 & 55.50 & 
    42.68 / 92.66 & 2.940 & 
    20.98 / 84.88 & 53.81 & 
    27.22 / 87.70 & 56.41 \\
    4RC~\cite{luo20264rc} & 
    \underline{85.53} / \underline{98.13} & \underline{86.53}  &
    37.79 / 89.97 & 71.97             &
    92.07 / 99.20 & \underline{1.380} &
    26.45 / 86.66 & \underline{54.14} &
    78.96 / 97.58 & 73.90             \\
    \midrule
    SM4RT & 
       \textbf{86.16} /    \textbf{98.36} & \textbf{89.89}    & 
    \underline{43.88} /    \textbf{91.56} & \underline{74.86} &   
       \textbf{93.30} /    \textbf{99.32} & \textbf{1.153}   & 
       \textbf{27.91} /    \textbf{86.76} & \textbf{54.31}    & 
    \underline{81.45} / \underline{97.88} & \textbf{78.50}    \\
    \bottomrule
  \end{tabular}
}
\vspace{-5mm}
\end{table*}

\section{Experiments}

\subsection{Implementation Details}

Our base model architecture follows Depth Anything 3~\cite{lin2026depth}.
We initialize \textbf{SM4RT} with pre-trained weights for the \textcircled{1} DINOv2 Backbone, \textcircled{2} DepthHead, and \textcircled{3} CameraHead.
The number of motion tokens $N$ is set to 10. Overall, the model has 1.38 billion parameters, with motion-specific modules accounting for 0.17 billion. 
Training was conducted on 8 NVIDIA RTX5880 GPUs for 5 days.
We employed the AdamW optimizer and a hybrid schedule of linear warm-up (first 0.5 epochs) followed by cosine decay, with learning rate adjusted within [1e-7,1e-4].
For Lie group exponential mapping, we use PyTorchSE3~\cite{pytorchse3}.

\subsection{Training Details}

We curated a collection of four dynamic datasets with motion annotations: Kubric (Movi-F)~\cite{greff2022kubric}, Stereo4D~\cite{jin2024stereo4d}, PointOdyssey~\cite{zheng2023pointodyssey}, and DynamicReplica~\cite{karaev2023dynamicstereo}. Additionally, for fine-tuning, we include datasets with scene geometry annotations: ARKitScenes~\cite{baruch2021arkitscenes}, HyperSIM~\cite{roberts2021hypersim}, ScanNet++~\cite{yeshwanth2023scannetpp}, VKitti2~\cite{cabon2020virtual}, and Waymo~\cite{sun2020scalability}. 

For Kubric, we retrieve \textcircled{1} world-coordinate positions and transform them to the first-frame camera coordinate system using the Procrustes algorithm, together with \textcircled{2} background segmentation. For Stereo4D, we filter high-quality data with denser annotations ($>10K$ valid points per scene) to avoid under-supervision. 
Point coordinates are transformed and normalized in world coordinates.
The number of input frames varies from 12 to 24. 
The scene scale is normalized such that the mean Euclidean distance from all valid 3D points to the origin equals 1.

We employ a multi-stage training strategy. 
In stage I, we pretrain the model on Kubric for 100 epochs, which is necessary for learning base assignments. 
In stage II, we include all motion datasets for 150 epochs. 
Each epoch contains 8000 episodes. 
In stage III, we fine-tune depth and camera prediction on static and dynamic scenes for 25 epochs. 

\begin{table}[t]
\caption{\textbf{TapVid3D benchmark results on the minval split.} We report 3D Average Jaccard (AJ) and Average Percent of Points within Delta (APD) to evaluate tracking performance. * indicates that the results are from~\cite{liu2025traceanythingrepresentingvideo}.}
\label{tab:tapvid3d}
\vspace{-3mm}
\centering
\resizebox{\linewidth}{!}{
\begin{tabular}{l|rr|rr|rr}
\toprule
Task & \multicolumn{6}{c}{\textbf{TapVid-3D Tracking}} \\
\midrule
Dataset & \multicolumn{2}{c|}{\textbf{ADT}} & \multicolumn{2}{c|}{\textbf{DriveTrack}} & \multicolumn{2}{c}{\textbf{PStudio}} \\
\midrule
Model & AJ & APD & AJ & APD & AJ & APD \\
\midrule
SpatialTracker~\cite{SpatialTracker} &
9.2 & 15.1 & 5.8 & 10.2 & 9.8 & 17.7 \\
DELTAv1~\cite{ngo2025delta} &
20.3 & 20.4 & 18.0 & 25.2 & \textbf{18.5} & 26.1 \\
DELTAv2~\cite{ngo2025DELTAv2} &
20.5 & 20.9 & 19.1 & 26.2 & \underline{17.9} & \underline{26.2} \\
SpatialTrackerV2~\cite{xiao2025spatialtrackerv2} &
18.1 & 28.2 & 14.2 & 20.9 & 16.9 & \textbf{26.3} \\
\midrule
St4RTrack~\cite{feng2025st4rtrack}* &
13.4 & 15.2 & 7.4 & 8.5 & 6.9 & 7.2 \\
TraceAnything~\cite{liu2025traceanythingrepresentingvideo}* &
15.6 & 20.5 & 9.6 & 15.5 & 10.8 & 16.3 \\
D4RT~\cite{zhang2026d4rt,li2026opend4rt} &
24.3 & 31.5 & 14.7 & 19.3 & 17.5 & 23.5 \\
V-DPM~\cite{sucar2026v} &
28.3 & 35.3 & \underline{23.1} & \underline{28.6} & 17.6 & 25.5 \\
4RC~\cite{luo20264rc} &
\underline{30.2} & \underline{38.5} & \underline{23.1} & 28.3 & 16.2 & 22.8 \\
\midrule
SM4RT &
\textbf{32.5} & \textbf{40.6} & \textbf{24.3} & \textbf{30.1} & 16.7 & 23.6 \\
\bottomrule
\end{tabular}
}
\vspace{-5mm}
\end{table}

\begin{table*}[t]
\centering
\caption{\textbf{Performance on world-coordinate tracking.} We report Average Percent of Points within Delta (APD) and End-Point Error (EPE), where higher APD and lower EPE indicate better performance. Numbers are presented in fractions of All / Dynamic. Predictions are aligned to the ground truth using Global Alignment over 64 frames.}
\vspace{-3mm}
\label{tab:worldtrack}
\resizebox{\linewidth}{!}{
\begin{tabular}{l|cc|cc|cc|cc}
\toprule
Task & \multicolumn{8}{c}{\textbf{World-Coordinate Tracking}} \\
\midrule
Dataset &
\multicolumn{2}{c|}{\textbf{ADT}} & \multicolumn{2}{c|}{\textbf{DS}} &
\multicolumn{2}{c|}{\textbf{PO}} & \multicolumn{2}{c}{\textbf{PStudio}} \\
\midrule
Model & APD & EPE & APD & EPE & APD & EPE & APD & EPE \\
\midrule
MonST3R~\cite{zhang2025monst3r} &
74.35 / 67.92 & 0.2721 / 0.1578 &
58.06 / 51.86 & 0.4387 / 0.5313 &
33.47 / 39.36 & 0.9021 / 0.6452 &
51.32 / 51.32 & 0.4568 / 0.4568
\\
SpatialTracker~\cite{SpatialTracker} &
45.65 / 67.65 & 0.8530 / 0.1628 &
54.85 / 58.65 & 0.9274 / 1.0828 &
38.54 / 51.20 & 0.7499 / 0.4695 &
62.59 / 62.59 & 0.3094 / 0.3094
\\
St4RTrack~\cite{feng2025st4rtrack} &
76.00 / \underline{75.34} & 0.2680 / 0.1212 &
73.74 / 68.13 & 0.2682 / 0.2961 &
67.94 / 68.71 & 0.3140 / 0.2970 &
69.67 / 69.67 & 0.2637 / 0.2637
\\
TraceAnything~\cite{liu2025traceanythingrepresentingvideo} &
76.08 / 73.05 & 0.2469 / 0.1268 &
60.66 / 61.23 & 0.5756 / 0.4422 &
39.84 / 47.39 & 1.0595 / 0.7224 &
71.32 / 71.32 & 0.2727 / 0.2727
\\
Any4D~\cite{karhade2025any4d} &
58.65 / 70.66 & 0.4048 / 0.1313 &
79.08 / 68.81 & 0.2203 / 0.3112 &
66.20 / 66.55 & 0.3741 / 0.3991 &
64.44 / 64.44 & 0.2977 / 0.2977
\\
VDPM~\cite{sucar2026v} &
\underline{85.96} / 72.10 & 0.1666 / 0.1359 &
74.27 / \textbf{77.66} & 0.2548 / \textbf{0.2123} &
\underline{81.28} / \textbf{81.85} & \textbf{0.1899} / \textbf{0.1749} &
\underline{78.73} / \underline{78.73} & {0.1812} / {0.1812}
\\
4RC~\cite{luo20264rc} &
83.22 / 72.27 & 0.1905 / 0.1422 &
\underline{82.17} / 73.92 & \underline{0.1946} / \underline{0.2747} &
79.35 / \underline{75.94} & 0.2762 / 0.3517 &
70.91 / 70.91 & 0.2443 / 0.2433
\\
D4RT~\cite{zhang2026d4rt,li2026opend4rt} &
72.20 / 73.25 & 0.2758 / 0.3199 &
72.48 / \underline{74.88} & 0.2959 / 0.2755 &
67.99 / 74.25 & 0.3178 / \underline{0.2593} &
\textbf{79.60} / \textbf{79.60} & \textbf{0.1753} / \textbf{0.1753}
\\
\midrule
SM4RT &
\textbf{88.56} / \textbf{80.68} & \textbf{0.1339} / \underline{0.0935} &
\textbf{82.61} / 71.62 & \textbf{0.1918} / 0.2912 &
\textbf{81.62} / 73.68 & \underline{0.2578} / 0.3808 &
78.69 / 78.69 & \underline{0.1790} / \underline{0.1790}
\\
\textcolor{gray}{SM4RT (Adaptive)} & 
\textcolor{gray}{88.32 / 78.99} & \textcolor{gray}{0.1363 / 0.0995} & 
\textcolor{gray}{81.95 / 69.79} & \textcolor{gray}{0.1970 / 0.3051} & 
\textcolor{gray}{80.77 / 71.76} & \textcolor{gray}{0.2617 / 0.3897} & 
\textcolor{gray}{78.16 / 78.16} & \textcolor{gray}{0.1883 / 0.1883} 
\\
\bottomrule
\end{tabular}
}
\vspace{-4mm}
\end{table*}

\begin{figure*}[t]
    \centering
    \includegraphics[width=\linewidth]{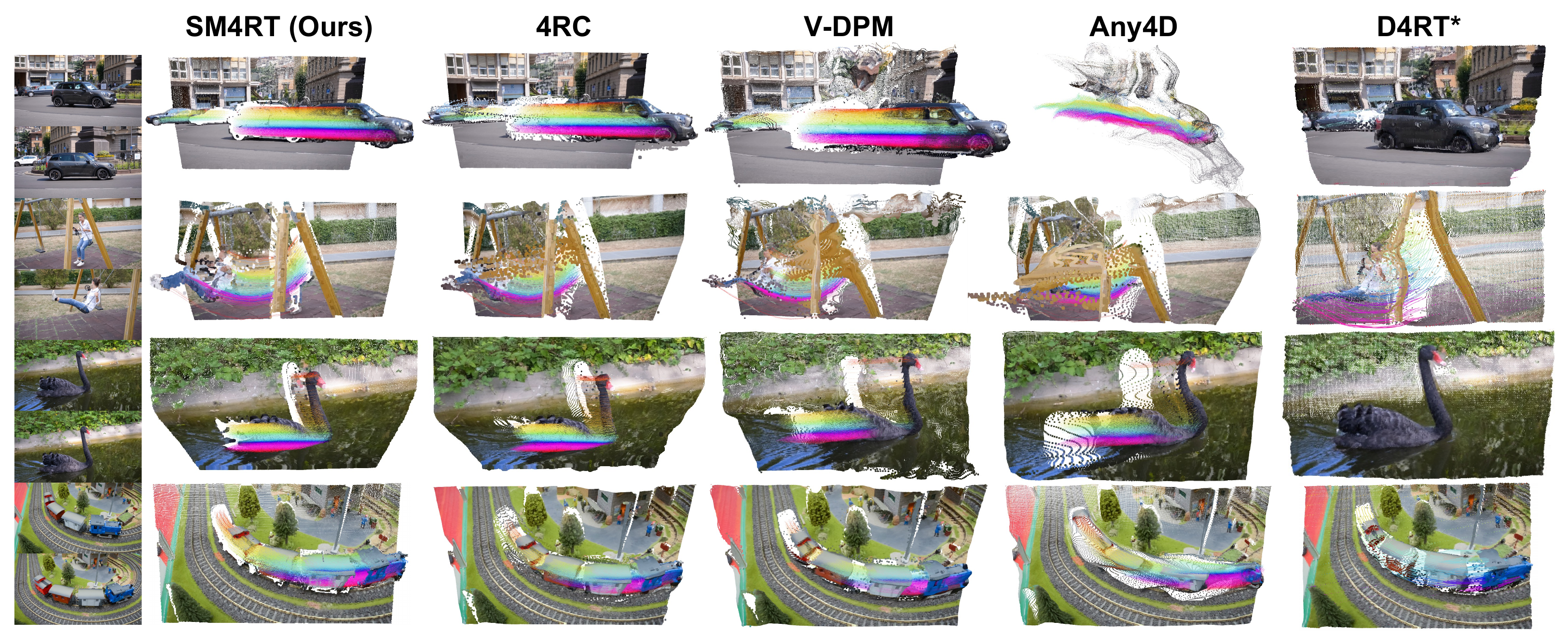}
    \vspace{-8mm}
    \caption{\textbf{Video track visualization.} We draw foreground tracklines and terminal positions of initial points. 
    We use the reproduced version of D4RT~\cite{li2026opend4rt} and sample input queries at a quarter resolution, while the other methods take dense input at resolution $294\times 518$.}
    \label{fig:Trackline}
    \vspace{-6mm}
\end{figure*}

\subsection{3D Geometry Reconstruction}

\textbf{Video Depth Estimation.}
\Cref{tab:depth} reports the video depth estimation performance following the protocol of MonST3R~\cite{zhang2025monst3r}. Our method, SM4RT, achieves competitive results across all benchmarks, reaching optimal or suboptimal in most datasets. These results demonstrate that SM4RT maintains strong depth estimation accuracy across diverse indoor and outdoor scenes.

\textbf{3D Reconstruction.}
\Cref{tab:DepthAnythingV3recon} presents the 3D reconstruction results on DA3Bench~\cite{lin2026depth}, evaluating both pose estimation (Auc3 / Auc30) and unposed reconstruction quality. SM4RT achieves the best overall performance across all five benchmarks. 
SM4RT consistently improves reconstruction quality across all datasets while maintaining comparable pose estimation performance, validating the effectiveness of our method for joint depth and pose estimation in 3D reconstruction pipelines.

\begin{figure*}[t]
    \centering
    \includegraphics[width=\linewidth]{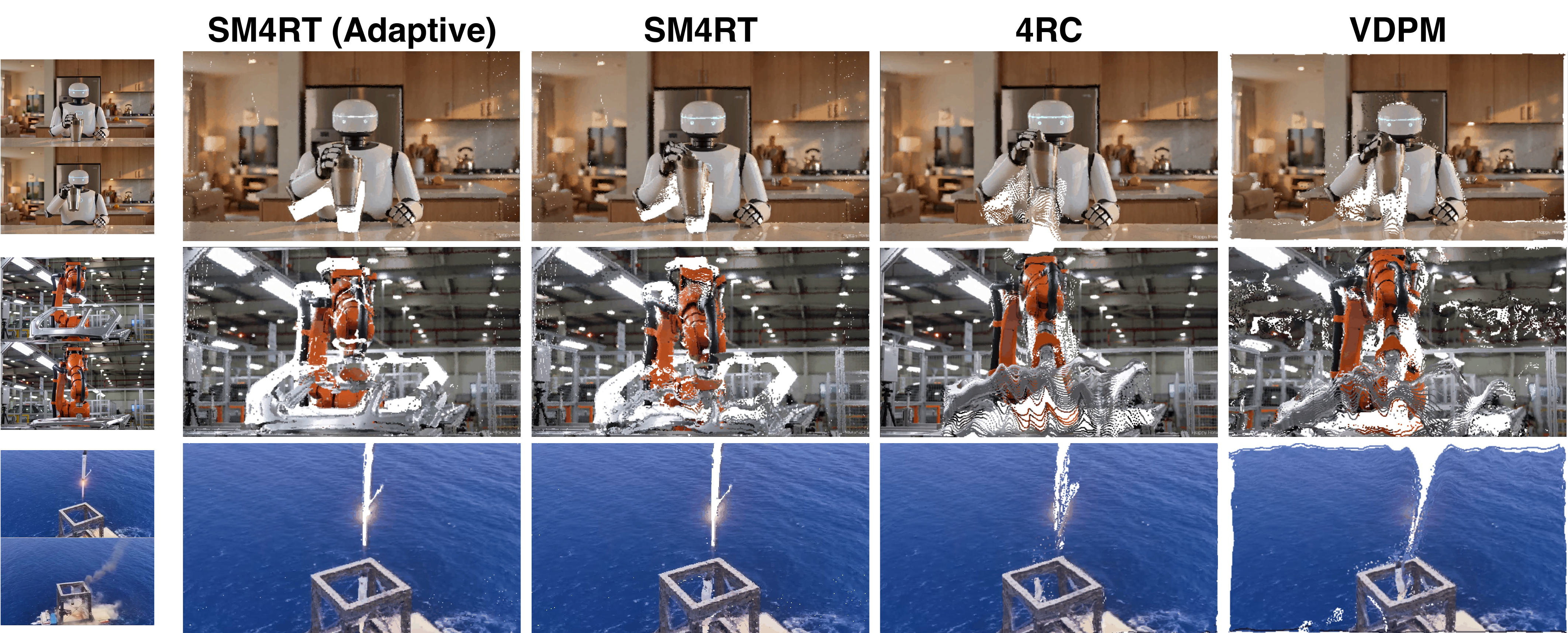}
    \vspace{-7mm}
    \caption{\textbf{Video motion visualization.} Instead of indirect tracklines, we present the rendered, unfiltered 2D projection of source-frame points in the target frame, with the reference source (top) and target (bottom) frames shown on the left. \textit{SM4RT} indicates that the motion is directly obtained from the output, while \textit{SM4RT (Adaptive)} automatically groups similar regions to obtain more structured motion.}
    \label{fig:MotionVisualization}
    \vspace{-3mm}
\end{figure*}

\begin{table*}[t]
\centering
    \caption{\textbf{Entity deformation and dense tracking on Kubric-Rigid (24 Frames).}}
    \label{tab:deformity}
  \begin{threeparttable}
    \vspace{-3mm}
 \setlength{\tabcolsep}{12pt}
\begin{tabular}{c|l|cc|c|cc}
\toprule
\multirow{2}{*}{Type} & \multirow{2}{*}{Model} & \multicolumn{2}{c|}{\textbf{Dynamic}} & \multicolumn{1}{c|}{\textbf{Initial}} & \multicolumn{2}{c}{\textbf{Tracking}} \\
\cmidrule{3-7}
 &  & VM$^{\textit{fg}}$ ($\downarrow$) & VM$^{\textit{bg}}$ ($\downarrow$)& VM$^{\textit{fg}}$ ($\downarrow$) & AJ ($\uparrow$) & APD ($\uparrow$) \\
\midrule
\multirow{2}{*}{Camera}
& DELTAv1~\cite{ngo2025delta}   & 
0.7137 / 0.6795 & 0.3223 & 0.4686 & 46.18 & 55.27
\\
& DELTAv2~\cite{ngo2025DELTAv2} & 
0.7503 / 0.7052 & 0.3405 & 0.4627 & 46.08 & 55.17
\\
\midrule
\multirow{6}{*}{World}
& SpatialTrackerV2~\cite{xiao2025spatialtrackerv2} &
0.3291 / 0.2821 & N/A & 0.7437 & N/A & N/A \\
& St4RTrack~\cite{feng2025st4rtrack}&
1.1597 / 0.9441 & 0.6971 & 1.0008 & 23.91 & 30.52
\\
& TraceAnything~\cite{liu2025traceanythingrepresentingvideo} &
\underline{0.1887} / 0.1761 & 0.1210 & 1.0081 & 76.33 & 85.61
\\
& Any4D~\cite{karhade2025any4d}     &
0.5053 / 0.4056 & 0.1385 & 0.6530 & 68.49 & 79.78
\\
& VDPM~\cite{sucar2026v}            &
0.1952 / \underline{0.1710} & 0.1811 & \textbf{0.2322} & 59.26 & 70.93
\\
& 4RC~\cite{luo20264rc} & 
0.2930 / 0.2503 & 0.2383 & 0.2520 & 82.25 & \underline{89.62}
\\
\midrule
\multirow{2}{*}{World}
& SM4RT & 
{0.1905} / {0.1876} & \underline{0.0122} & \underline{0.2452} & \textbf{84.37} & \textbf{90.62}
\\
& SM4RT (Adaptive) &
\textbf{0.0254} / \textbf{0.0191} & \textbf{0.0119} & {0.2499} & \underline{82.54} & {89.56}
\\
\bottomrule
\end{tabular}
  \end{threeparttable}
    \vspace{-6mm}
\end{table*}

\subsection{3D Tracking}

\textbf{Evaluations.}
We evaluate SM4RT under two standard 3D tracking settings. For TapVid3D~\cite{koppula2024tapvidd}, we follow the official minval split and report 3D Average Jaccard (AJ) and Average Percent of Points within Delta (APD) to measure tracking performance. As shown in \Cref{tab:tapvid3d}, SM4RT achieves the best results on ADT and DriveTrack across both metrics, outperforming prior methods such as V-DPM and 4RC. On PStudio, SM4RT remains competitive, although the best performance is achieved by other baselines.

We further evaluate world-coordinate tracking following the WorldTrack~\cite{feng2025st4rtrack} setting. In this protocol, predicted 3D point positions are aligned to the ground truth using Global Alignment over 64 frames. We report APD and End-Point Error (EPE), where higher APD and lower EPE indicate better tracking accuracy. Since D4RT~\cite{zhang2026d4rt} is not open-source, we adapt OpenD4RT~\cite{li2026opend4rt} to report its results under the same setting. As shown in \Cref{tab:worldtrack}, SM4RT achieves the best performance on ADT and strong results on DS and PO, showing accurate and stable 3D motion.

\textbf{Visualizations.}
In~\Cref{fig:Trackline}, we present visual comparisons of video tracklines on foreground objects. Compared with 4RC~\cite{luo20264rc}, V-DPM~\cite{sucar2026v}, Any4D~\cite{karhade2025any4d}, and D4RT~\cite{zhang2026d4rt,li2026opend4rt}, SM4RT produces more coherent tracklines that better adhere to object structure. While other methods suffer from scattered points or structural collapse, our method preserves moving-object geometry.

In \Cref{fig:MotionVisualization}, instead of showing indirect tracklines, we render the unfiltered 2D projection of source-frame points in the target frame, with the reference source and target frames shown on the left. SM4RT visualizes the motion directly predicted by the model, while SM4RT (Adaptive) applies adaptive grouped averaging according to the motion base assignment. Compared with 4RC~\cite{luo20264rc} and V-DPM~\cite{sucar2026v}, SM4RT produces more coherent projected motion with fewer scattered points and better preservation of object structure, especially near motion boundaries.

\subsection{Rigid Entity Deformation Analysis}

Previous work on evaluating rigid-body motion focuses only on 2D video generation models~\cite{jain2026rigidbench}. We propose the first benchmark for 3D rigid-body motion reconstruction.
The fundamental challenge in rigid-body motion is object deformation, which cannot be directly captured by existing point-wise metrics~\cite{koppula2024tapvidd}.
We therefore evaluate structure preservation using the average von Mises deformation score (VM), which measures non-rigid distortion after affine alignment between source and target point clouds.
This metric complements tracking accuracy by testing whether a method preserves each moving object as a coherent rigid entity rather than only localizing individual points.
The full metric derivation and Kubric-Rigid-Eval protocol are provided in Appendix~\ref{app:deformation}.

We evaluate on Kubric-Rigid-Eval, which contains 147 rigid-object videos with ground-truth segmentation and world-centric motion.
To compare camera- and world-coordinate trackers fairly, all methods are aligned by foreground-object scale and evaluated with both deformation and tracking metrics~\cite{koppula2024tapvidd}.

\Cref{tab:deformity} shows that with raw outputs, our method achieves state-of-the-art tracking performance and leading structure-preserving performance. 
While causing limited tracking degradation, the proposed adaptive grouping strategy nearly eliminates deformation caused by dynamic motion. 
The tracking drop may be due to inconsistent base assignments for some objects. Such assignments preserve structure but deviate slightly more from the ground truth.

\subsection{Motion Base Visualization}

\begin{figure}[t]
    \centering
    \includegraphics[width=\linewidth]{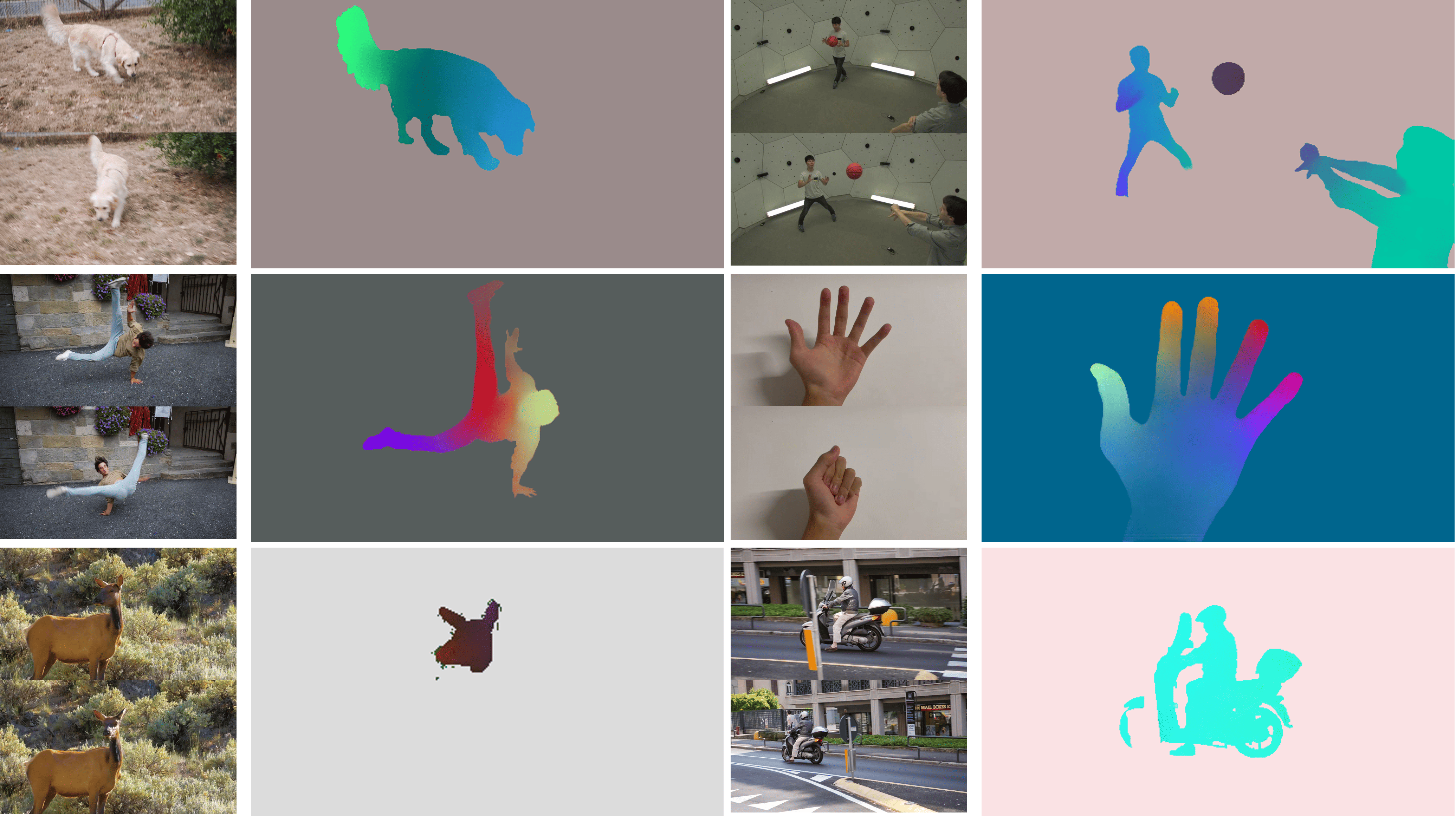}
    \vspace{-7mm}
    \caption{\textbf{Visualization of Motion Base Assignment.} We present the start and end frame of input video, and the corresponding PCA visualization of motion base assignment shared across the video.}
    \label{fig:moseg}
    \vspace{-3mm}
\end{figure}

Our method does not directly produce motion segmentation as in~\cite{SegAnyMo}. However, the weight assignment map can provide similar output. In fact, the weight assignment output is a structural segmentation that can be simplified into a dynamic mask. 
Some existing evaluations focus on segmenting entire motion entities (e.g., humans), making them unsuitable for evaluating our method because static parts of an entity would not be assigned as dynamic. As shown in~\Cref{fig:moseg}, our assignment map provides physically grounded, skeleton-aware structure, clearly delineating the articulated parts undergoing movement rather than treating the object as a monolithic dynamic blob. 
The weight assignment is shared by all frames and mounted on the reference frame. 

\subsection{Efficiency Analysis}

During inference, SM4RT shows superior computational efficiency while effectively extracting intricate motion structures. To comprehensively evaluate performance, we compare our method against several representative baselines under identical input resolutions. As summarized in \Cref{tab:inference_efficiency}, SM4RT achieves high inference speed and low GPU memory footprint, striking a strong balance between speed and resource consumption. Notably, the SM4RT (Adaptive) variant improves structural integrity at modest cost. 

It is important to note that the results of V-DPM are due to its fundamentally different paradigm, which incurs significantly higher computational costs. However, when evaluated on a per-step basis (Single Pass), its computational cost is comparable to that of the other baselines. 

\subsection{Ablation Study}

\begin{table}[t]
\caption{\textbf{Comparison of inference efficiency among different methods.} Results are measured under the same input resolution and frame number on BF16 precision averaged over the 147 samples of the Kubric evaluation set. Inference is conducted on a single NVIDIA RTX5880 (48G).}
    \vspace{-3mm}
\label{tab:inference_efficiency}
\resizebox{\linewidth}{!}{
\begin{tabular}{l|cccc}
\toprule
\textbf{Method} & 
\textbf{Time (s/sample)} & \textbf{TFLOPS} & \textbf{Memory (GB)} & \textbf{APD} \\
\midrule
DELTAv2 (iter1)~\cite{ngo2025DELTAv2} &  6.35 &  68.91 & 36.02 & 54.72 \\
DELTAv2 (iter4)~\cite{ngo2025DELTAv2} &  7.31 &  79.61 & 36.02 & 55.17 \\
DELTAv2 (iter6)~\cite{ngo2025DELTAv2} & 11.05 & 142.27 & 36.02 & 55.47 \\
\midrule
St4RTrack~\cite{feng2025st4rtrack}      
& 2.54 &  61.58 & 15.77 & 30.52 \\
TraceAnything~\cite{liu2025traceanythingrepresentingvideo}  
& 1.74 & 109.71 & 19.91 & 85.61 \\
Any4D~\cite{karhade2025any4d}    
& 4.53 & 103.95 & 16.88 & 79.78 \\
V-DPM (All Pass)~\cite{sucar2026v}
& 64.79 & 6467.91 & 29.78 & 70.93 \\
V-DPM (Single Pass)~\cite{sucar2026v}
& 4.30 & 438.00 & 24.22 & N/A \\
4RC~\cite{luo20264rc}
& 4.24 & 278.40 & 15.22 & 89.62 \\
\midrule
\textbf{SM4RT} & 2.86 & 177.74 & 15.72 & 90.62 \\
\textbf{SM4RT (Adaptive)} & 11.64 & 210.85 & 16.65 & 89.56 \\
\bottomrule
\end{tabular}
}
    \vspace{-5mm}
\end{table}

We investigated the impact of different backbones and intermediate layer injection strategies on model performance. 
As shown in Table~\ref{tab:ablation_study}, the results demonstrate that the DA3 backbone outperforms VGGT largely, achieving higher scores, highlighting the effective of our backbone design. 
Among the fusion strategies, the linear injection method yields the best performance, surpassing both MLP and Replace variants in both metrics. This suggests that a simple linear transformation is sufficient and effective for integrating intermediate features.

In~\Cref{tab:ablation_study_ft}, we compare different fine-tuning settings for final stage. 
``Rigid'' targets on structure-keeping using singularity loss, while ``Non-Rigid'' applies traditional point supervision on non-Rigid datasets, and ``All'' mix them together. 
On Tracking, the ``All" setting achieves the best overall performance, indicating that mixed-data finetuning effectively improves tracking accuracy. On Deformity, while the ``Non-Rigid" setting to some extend harness structural rigidity. In contrast, the ``All" setting provides a more balanced performance on both criterion.

\begin{table}[t]
    \centering
    \caption{\textbf{Ablation study on intermediate injection strategies.} 
    We compare different backbones (VGGT and DA3) and fusion strategies under the DA3 backbone, under APD metric.}
    \label{tab:ablation_study}
\vspace{-3mm}
\resizebox{\linewidth}{!}{
    \begin{tabular}{l|ccc}
        \toprule
        \textbf{Backbone} & \textbf{Injection Strategy} & \textbf{ADT} & \textbf{DS} \\
        \midrule
        VGGT & Linear & 80.79 / 77.73 & 72.48 / 71.39 \\
        \midrule
        DA3 & Linear & \textbf{88.56} / \textbf{80.68} & \textbf{82.61} / \textbf{71.62} \\
        DA3 & MLP & 87.65 / 78.28 & 81.05 / 69.26 \\
        DA3 & Replace & 85.00 / 78.51 & 80.58 / 69.14 \\
        \bottomrule
    \end{tabular}
}
\vspace{-3mm}
\end{table}

\begin{table}[t]
    \centering
    \caption{\textbf{Ablation study on data training setting.} 
    We compare tracking (APD) deformation (VM).}
    \label{tab:ablation_study_ft}
\vspace{-3mm}
\resizebox{\linewidth}{!}{
    \begin{tabular}{l|ccc}
        \toprule
        \textbf{Setting} & \textbf{ADT} & \textbf{DS} & \textbf{Kubric (VM)} \\
        \midrule
        Rigid & 
        87.76 / 79.74 & 81.39 / 69.46 & \textbf{0.1847} / \textbf{0.1203}  \\
        Non-Rigid & 
        86.01 / 81.45 &  82.38 / \textbf{71.69} & 0.2347 / 0.2186 \\
        All & 
        \textbf{88.56} / \textbf{80.68} & \textbf{82.61} / {71.62} & 0.1905 / 0.1876 \\
        \bottomrule
    \end{tabular}
}
\vspace{-6mm}
\end{table}

\section{Conclusion}

We have presented SM4RT, a framework that extends Geometry Foundation Models from static 3D reconstruction to structured 4D dynamic understanding.
The central contribution is a shift in the motion representation paradigm.
Rather than predicting independent point-wise displacements, SM4RT introduces Structure-of-Motion (SoM): scene dynamics are decomposed into a compact set of $N$ motion bases, each represented as a temporal sequence of 6D twists in $\mathfrak{se}(3)$.
Dense scene motion is then recovered by sparse, time-shared per-pixel assignment weights over these bases.
This reduces the effective motion degrees of freedom from $3HW$ to $6N$, enforces object-level kinematic coherence by construction, and yields a representation that is both physically interpretable and parsimonious.
Built on a pretrained GFM backbone with a parallel Motion Geometry Encoder and Decoder, SM4RT jointly infers 3D geometry, world-coordinate motion, and scene kinematic structure in a single forward pass from monocular RGB video.
It achieves advanced performance on motion reconstruction and uniquely enables dynamic motion structure perception.

\newpage

\appendix
    {\Large\bfseries Appendix}

\section{Rigid Entity Deformation Details}
\label{app:deformation}

We measure rigid-entity deformation with a von Mises deformation score (VM) derived from continuum mechanics~\cite[pp.~90--95]{deSaracibar2023nonlinear}.
For each rigid segment, we first align the source point cloud to the target point cloud with an affine transformation:
\begin{equation}
    (\bm{F}^*, \bm{c}^*) = \underset{\bm{F}, \bm{c}}{\operatorname{argmin}} \sum_{i=1} \left\| \bm{x}_{\textit{tgt}}^{(i)} - \left( \bm{F}\bm{x}_{\textit{src}}^{(i)} + \bm{c} \right) \right\|^2_2,
\end{equation}
where $\bm{F}^* = \bm{R}\bm{U}$ contains rotation $\bm{R}$ and stretch $\bm{U}$.
Since this estimates a relative transform, the resulting $\bm{F}$ is scale-invariant.
We then compute the right Cauchy-Green deformation tensor $\bm{C} = (\bm{F}^*)^\top \bm{F}^* = \bm{U}^\top \bm{U}$ and derive the Green-Lagrange strain tensor $\bm{E}$:
\begin{equation}
    \bm{E} = \frac{1}{2} \left( \bm{C} - \bm{I} \right) = \frac{1}{2} \left( \bm{U}^\top \bm{U} - \bm{I} \right),
\end{equation}
where $\bm{I}$ is the identity tensor.
The tensor $\bm{E}$ captures stretching and shearing components in the global transformation, enabling the assessment of structural deformation.
We compute the von Mises equivalent Green-Lagrange strain $\epsilon_{\text{vM}}$ from the deviatoric part of the strain tensor:
\begin{align}
    \epsilon_{\text{vM}} &= \sqrt{\frac{2}{3} \bm{E}' : \bm{E}'} = \sqrt{\frac{2}{3} \sum_{i,j} \bm{E}'_{ij} \bm{E}'_{ij}},
    \\
    \bm{E}' &= \bm{E} - \frac{1}{3}\operatorname{tr}(\bm{E})\bm{I}.
\end{align}
This scalar quantifies the intensity of non-rigid distortion.
Given video sequences of rigid-body scenes with ground-truth instance segmentations, we compute $\epsilon_{\text{vM}}$ weighted by the number of segment points and frames to obtain VM.

\textbf{Evaluation protocol.}
Kubric-Rigid-Eval contains 147 video samples of rigid objects with ground-truth segmentation and world-centric motion.
The evaluation reports \textcircled{1} initial foreground deformation, \textcircled{2} dynamic foreground deformation without and with edge filtering, and \textcircled{3} dynamic background deformation.
Dynamic object deformation measures how the target point cloud deforms from the source point cloud, while initial object deformation measures how well the initial position aligns with the ground-truth point cloud.
Dynamic deformation is scale-invariant because it is measured within each sample, while initial deformation is aligned to the initial ground truth by object-wise scale.
Together, these terms quantify deformation from complementary sources.

\section{Limitations}
The rigid-body assumption underlying SoM does not account for physical scene interactions such as collisions, contact dynamics, or fluid deformation, limiting the fidelity of future state prediction in highly complex scenarios. 
Additionally, the number of motion bases $N$ is fixed at inference time. An adaptive or hierarchical decomposition could better handle scenes with highly variable numbers of moving objects.
Therefore, SM4RT may also perform less reliably on highly deformable objects, leaving the balance between rigid and deformable motion as an open issue. 
We view SM4RT as establishing a foundation for structured 4D scene understanding and hope it inspires further work on physically grounded, interpretable dynamic representations.

\end{document}